%% file: main.tex
\pdfoutput=1

\documentclass[11pt]{article}

\usepackage[final]{acl}
\usepackage{ulem}
\usepackage{times}
\usepackage{floatrow}
\usepackage{latexsym}

\usepackage[T1]{fontenc}

\usepackage[utf8]{inputenc}

\usepackage{microtype}

\usepackage{inconsolata}

\usepackage{graphicx}
\usepackage{booktabs}
\usepackage{multirow}
\usepackage{tcolorbox}
\usepackage{afterpage}
\usepackage{booktabs}
\usepackage{subcaption}
\usepackage{placeins}
\usepackage{array, ragged2e}
\usepackage{tabularx}
\usepackage{url}
\title{AcT2I: Evaluating and Improving Action Depiction  in\\ Text-to-Image Models}

\author{
 \textbf{Vatsal Malaviya},
 \textbf{Agneet Chatterjee},
 \textbf{Maitreya Patel},
 \textbf{Yezhou Yang},
 \textbf{Chitta Baral}
\\
 Arizona State University
}

\begin{document}

\maketitle
\input{sec/0_abstract}
\FloatBarrier
\begin{figure}
    \centering
    \includegraphics[width=\linewidth]{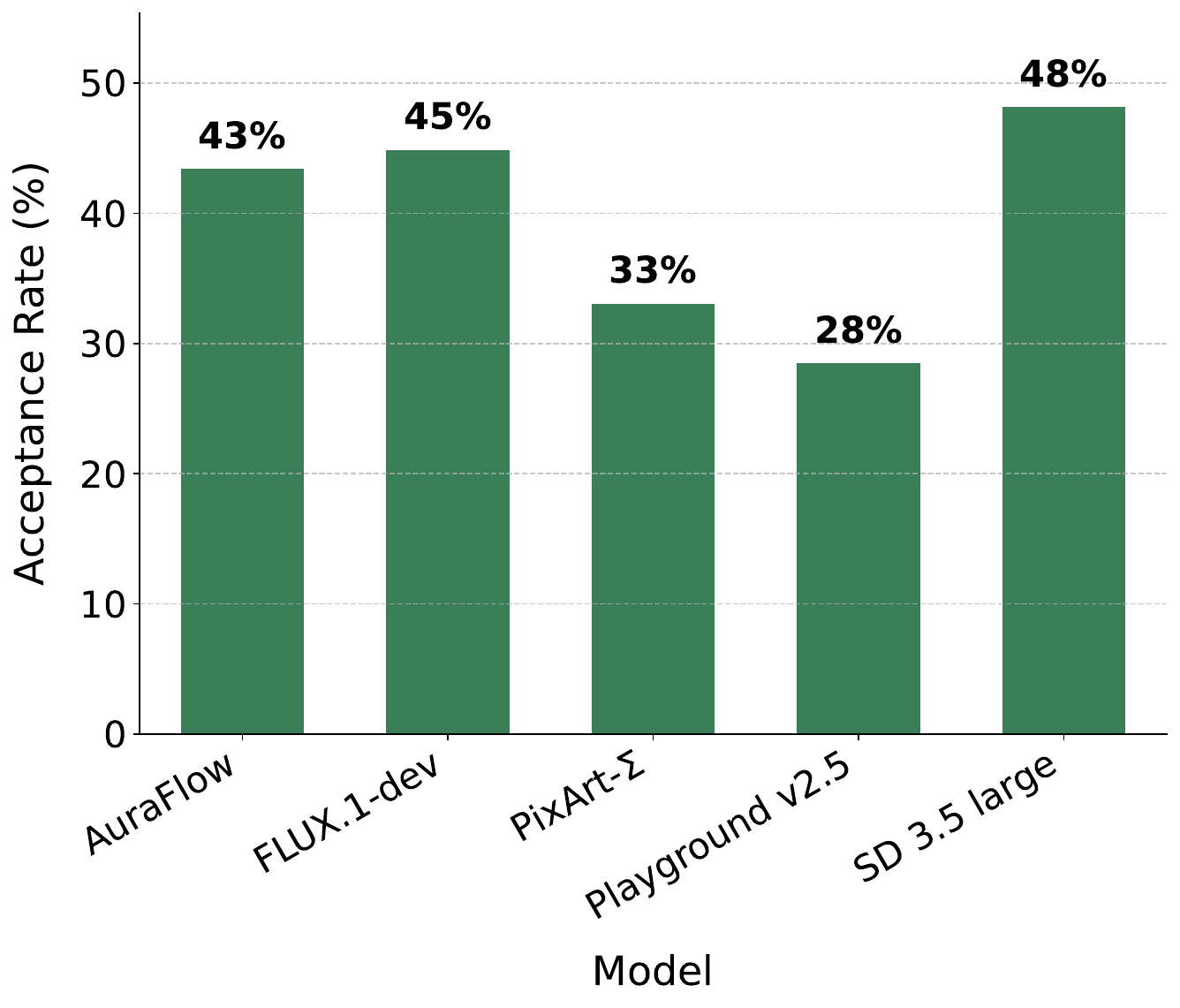}
    \caption{Action Depiction Performance of SOTA Text-to-Image Models. Each generated image was evaluated by three reviewers, resulting in a binary acceptance (yes/no). The acceptance rate represents the average agreement among reviewers on each model's acceptability of action depiction.} 
    \label{fig:accept-rate}
\end{figure}

\afterpage{%
\begin{figure*}[!h]
    \centering
    \includegraphics[width=\linewidth]{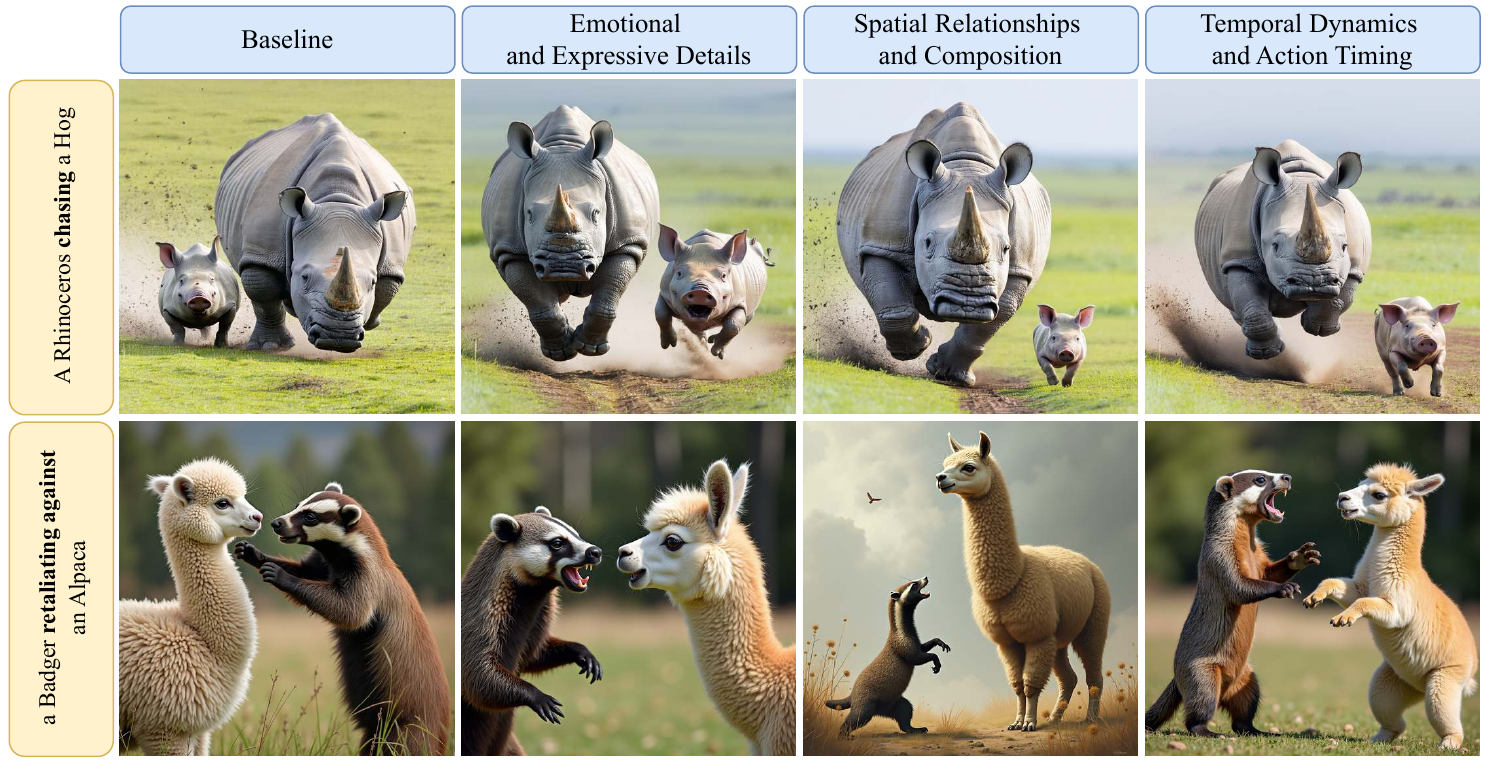}
    \caption{LLM-Transformed prompts unlock richer and accurate action depiction. Compared to baseline prompts that lack dense details, LLM-guided transformation into Emotional, Spatial, and Temporal dimensions generates images exhibiting more compelling action dynamics, finer expressive details, and improved subject placement accuracy. (See Appendix~\ref{sec:qualitative-grid} for an extended qualitative analysis.)}
    \label{fig:teaser}
\end{figure*}
}
\input{sec/1_intro}
\input{sec/2_related_works}
\input{sec/3_experiments}
\input{sec/4_conclusion}
\input{sec/5_limitations}
\bibliography{main}
\newpage
\input{sec/z_appendix}
\end{document}

%% file: sec/0_abstract.tex
\begin{abstract} 

Text-to-Image (T2I) models have recently achieved remarkable success in generating images from textual descriptions. However, challenges still persist in accurately rendering complex scenes where actions and interactions form the primary semantic focus. Our key observation in this work is that T2I models frequently struggle to capture nuanced and often implicit attributes inherent in action depiction, leading to generating images that lack key contextual details. To enable systematic evaluation, we introduce AcT2I, a benchmark designed to evaluate the performance of T2I models in generating images from action-centric prompts. We experimentally validate that leading T2I models do not fare well on AcT2I. We further hypothesize that this shortcoming arises from the incomplete representation of the inherent attributes and contextual dependencies in the training corpora of existing T2I models. We build upon this by developing a training-free, knowledge distillation technique utilizing Large Language Models to address this limitation. Specifically, we enhance prompts by incorporating dense information across three dimensions, observing that injecting prompts with temporal details significantly improves image generation accuracy, with our best model achieving an increase of  72\%. Our findings highlight the limitations of current T2I methods in generating images that require complex reasoning and demonstrate that integrating linguistic knowledge in a systematic way can notably advance the generation of nuanced and contextually accurate images. Project Page : \url{https://vatsal-malaviya.github.io/AcT2I/}

\end{abstract}

%% file: sec/1_intro.tex
\section{Introduction}
\label{sec:intro}
Text-to-Image (T2I) models have advanced rapidly, evolving from simple image generation systems to producing intricate, photorealistic scenes \cite{karras2019style, rombach2022high, esser2024scaling}. There has been a consistent growth in the performance of T2I models in their ability to perform compositional tasks such as object placement and attribute binding, as evidenced by their performance on benchmarks such as T2I-CompBench \cite{huang2023t2i} and GenEval \cite{ghosh2024geneval}.

However, these benchmarks aim to capture \textit{1-hop } capabilities of T2I models, i.e. evaluating abilities that does not require nuanced reasoning while generating an image. For example, to generate a "\textit{blue apple}", the model has to follow 2 steps: \textbf{1}. generating an apple followed by, \textbf{2}. coloring it blue. While being important, these setups fail to accurately evaluate T2I models in generating complex scenarios which require multiple iterations of reasoning. Furthermore, with the rapid development of state of the art T2I models, it is imperative to develop stringent benchmarks.

To address this gap, we develop the \textbf{AcT2I} benchmark, to evaluate Text-to-Image models in their ability to generate images of action-centric scenarios. To the best of our knowledge, action depiction has not been studied for T2I models. In this work, we aim to define this problem statement, benchmark existing models and develop baseline methods, with the ultimate goal of aligning T2I models with human interpretation of actions, which are inherently complex. For example, depicting “\textit{a viper coiling around a duck}” involves more than just object placement and spatial composition; it requires understanding of relative object proportions, temporal dynamics and appropriate emotional expressions.

We develop the \textbf{AcT2I} evaluation benchmark by sampling a total of 20 actions from the Animal Kingdom \cite{ng2022animalkingdomlargediverse} dataset, covering 100 animals, developing a total of 125 prompts. With this evaluation suite, we perform a comprehensive human evaluation, across 5 state of the art T2I models. Our key finding is that - existing T2I models struggle to generate images that accurately depict realistic actions based on textual prompts. These models tend to overfit to conventional actions associated with specific animals and fail to capture the nuanced details necessary to convey a given action effectively. Our next observation is that providing dense information in the text prompt improves performance in action generation by a significant margin. Therefore, we propose a test-time 
Large Language Model (LLM) \cite{touvron2023llama, openai2024gpt4technicalreport} guided knowledge distillation pipeline that enhances the text prompt across multiple dimensions, leading to up to \textit{3x} gains in performance, as shown in Figure \ref{fig:teaser}.

To summarize, our contributions are as follows: 
\begin{itemize} 
    \item We develop the \textbf{AcT2I} benchmark to evaluate the ability of T2I models to generate images from textual prompts that describe actions. We evaluate a total of 25 actions across 100 animals, sampled from the Animal Kingdom \cite{ng2022animalkingdomlargediverse} dataset.
    \item Our findings, based on an extensive evaluation of 5 state-of-the-art T2I models, reveal a significant limitation in their ability to generate accurate and realistic action depictions.  
    \item We develop a training-free LLM-guided knowledge distillation technique that injects dense descriptions into prompts across 3 dimensions - spatial, temporal and emotional; and find large gains in performance, such as a 73\% improvement in the performance of Stable Diffusion 3.5 Large.
\end{itemize}

%% file: sec/2_related_works.tex
\section{Related Works}
\label{sec:related_work}

\subsection{Text-to-Image Models}
Early Text-to-Image (T2I) models focused on generating simple, often low-resolution images directly from textual prompts. This changed with Stable Diffusion \cite{rombach2022high}, which pioneered latent-space processing using VQGAN \cite{esser2021taming}, enabling scalable, high-fidelity image generation. Subsequent efforts, such as Latent Consistency Models \cite{luo2023latent} and InstaFlow \cite{liu2023instaflow}, have further optimized aspects like generation speed and image quality. Recently, models including FLUX.1-dev \cite{flux2024} and Stable Diffusion 3 \cite{esser2024scaling} have pushed the boundaries of compositional accuracy and realism. However, despite these advancements, current T2I models often struggle with capturing intricate relationships and complex interactions, motivating the need for additional techniques to enhance semantic understanding.

\subsection{Knowledge Distillation from Large Language Models}
Descriptive captions have proven to improve image generation in text-to-image models \cite{betker2023improving}.  Knowledge Distillation (KD) provides a pathway to improve T2I models by transferring semantic and contextual knowledge from Large Language Models (LLMs) without additional full-scale retraining. For instance, KD-DLGAN \cite{cui2023kd} leverages generative distillation to enhance image diversity and quality even under limited data conditions. Beyond improving visual realism, KD can bridge the gap between textual semantics and visual representations, facilitating tasks like visual question answering and image captioning. Augmenting models like CLIP \cite{radford2021learning} with LLM-derived knowledge has shown promise in improving vision-language alignment \cite{dai2022enabling}. Nevertheless, current KD approaches often assume static relationships \cite{feng2024layoutgpt, wu2024self, datta2023prompt, zhong2023adapter} and lack the ability to handle dynamic, action-centric scenarios. This limitation underscores the need to adapt KD techniques for more sophisticated tasks where temporal and relational dynamics play a critical role.

\subsection{Relational Understanding in Generative Models}
While T2I models now excel at producing photorealistic images, they remain limited in their capacity to generate coherent relational scenes. Existing SOTA models—such as Stable Diffusion 3.5 Large, DALL-E 3 \cite{betker2023improving}, and FLUX.1-dev—often fail to correctly depict scenarios like “a cat chasing a mouse under a table,” yielding images that lack logical spatial arrangements or contextual correctness \cite{chatterjee2024getting, conwell2022testing, lian2023llm}. The core challenge is that these models typically do not capture fine-grained relational cues, making it difficult to represent dynamic interactions or hierarchical relationships \cite{fu2024commonsense}. Benchmarks like the Textual-Visual Logic Challenge \cite{10.1007/978-3-031-72652-1_19} highlight these shortcomings, focusing on compositional and logical consistency rather than isolated attributes.

In this work, we develop a benchmark that requires relational reasoning of a complex form -- generating the correct action between two entities. We empirically establish that this is indeed a hard problem for existing T2I models and develop a simple baseline method to improve upon this existing shortcoming.

%% file: sec/3_experiments.tex
\section{Benchmarking T2I Models on AcT2I}
\label{sec:experiments}

\subsection{Experimental Setup}
We evaluate 5 T2I models — Stable Diffusion 3.5 Large \cite{esser2024scaling}, AuraFlow\footnote{https://huggingface.co/fal/AuraFlow-v0.2}, FLUX.1-dev \cite{flux2024}, Playground v2.5 \cite{li2024playgroundv25insightsenhancing}, and PixArt-$\Sigma$ \cite{chen2024pixartsigmaweaktostrongtrainingdiffusion};  with these models varying across their pre-training data, diffusion architecture and text encoders. We sample 4 images per prompt to maintain consistency. All image generations were performed using publicly  available model checkpoints with parameter settings, unless otherwise specified.  All experiments were run on a NVIDIA A100 GPUs.

\subsection{Prompt Generation}
\label{subsec:prompt_generation}
All our prompts consist of 2 entities and 1 action relationship; example prompts are enumerated in Table \ref{tab:prompt_examples}. Our entities and actions are sourced from the Animal Kingdom Dataset \cite{ng2022animalkingdomlargediverse}. We choose this setup because it enables to evaluate animals from diverse taxonomies (\textit{for example}, mammals and reptiles) and sample actions of varying kinds. Furthermore, unlike human image generation (More details in Appendix \ref{sec:generalization}), T2I models do not exhibit issues such as disfigurement when generating animal images. This allows us to focus exclusively on evaluating their ability to depict actions accurately.

\input{tables/prompt_examples}

We cover 25 actions, generating 5 prompts/action, each instantiated with a unique animal-animal combination, leading to a total of 125 prompts. Our prompts cover both actions naturally associated with a given animal and those that are less typical, ensuring coverage of both in-distribution and out-of-distribution scenarios. Overall, our benchmark ensures a broad coverage of action types, evaluating the models’ compositional reasoning, and facilitate a meaningful assessment of their ability to depict nuanced animal interactions.

To ensure diversity and coverage, each of our 125 prompts was annotated along four orthogonal axes: interaction rarity (frequent, rare, very rare), emotional valence (aggressive, defensive, affiliative, communicative), spatial topology (proximal contact, pursuit/avoidance, distal), and temporal extent (instantaneous vs extended). The resulting distribution (Frequent 31\%, Rare 31\%, Very rare 38
\%; Aggressive 56\%, Defensive 20\%, Affiliative 12\%, Communicative 12\%; etc.) shows that no axis is dominated by a single category. We deliberately overrepresented rare and extended interactions (e.g., “a goose retaliating against a giraffe”) to push models beyond memorized patterns. A Shannon entropy analysis across all axes (> 0.82) further confirms dataset balance. Detailed per-axis statistics and performance breakdowns are provided in Appendix \ref{sec:dataset_coverage}.

\subsection{Annotation Setup}
A total of 25 annotators, hired via Amazon Mechanical Turk where each image was independently rated by 3 annotators. Each annotator was instructed to answer "Yes" or "No" to the question: "Does the image accurately depict the action described in the prompt?". We define the acceptance rate as the proportion of images receiving a “Yes” from a majority of annotators. This binary evaluation helps isolate whether models can convey the intended action rather than focusing on nuanced aesthetic qualities. More details are presented in Appendix \ref{subsec:annot_details}.

\subsection{Benchmarking Results}

\textbf{Overall Performance}: As shown in Figure \ref{fig:accept-rate}, Stable Diffusion 3.5 Large achieves the highest acceptance rate (48\%), followed by FLUX.1-dev (45\%) and AuraFlow (44\%). In contrast, PixArt-$\Sigma$ and Playground v2.5 lag behind at 29\% and 27\%, respectively. These results indicate a considerable performance gap, with no model surpassing a 50\% acceptance rate across challenging action-centric prompts, indicating that none of the models get majority of the images correct. 

\begin{figure*}[!ht]
    \centering
    \includegraphics[width=\linewidth]{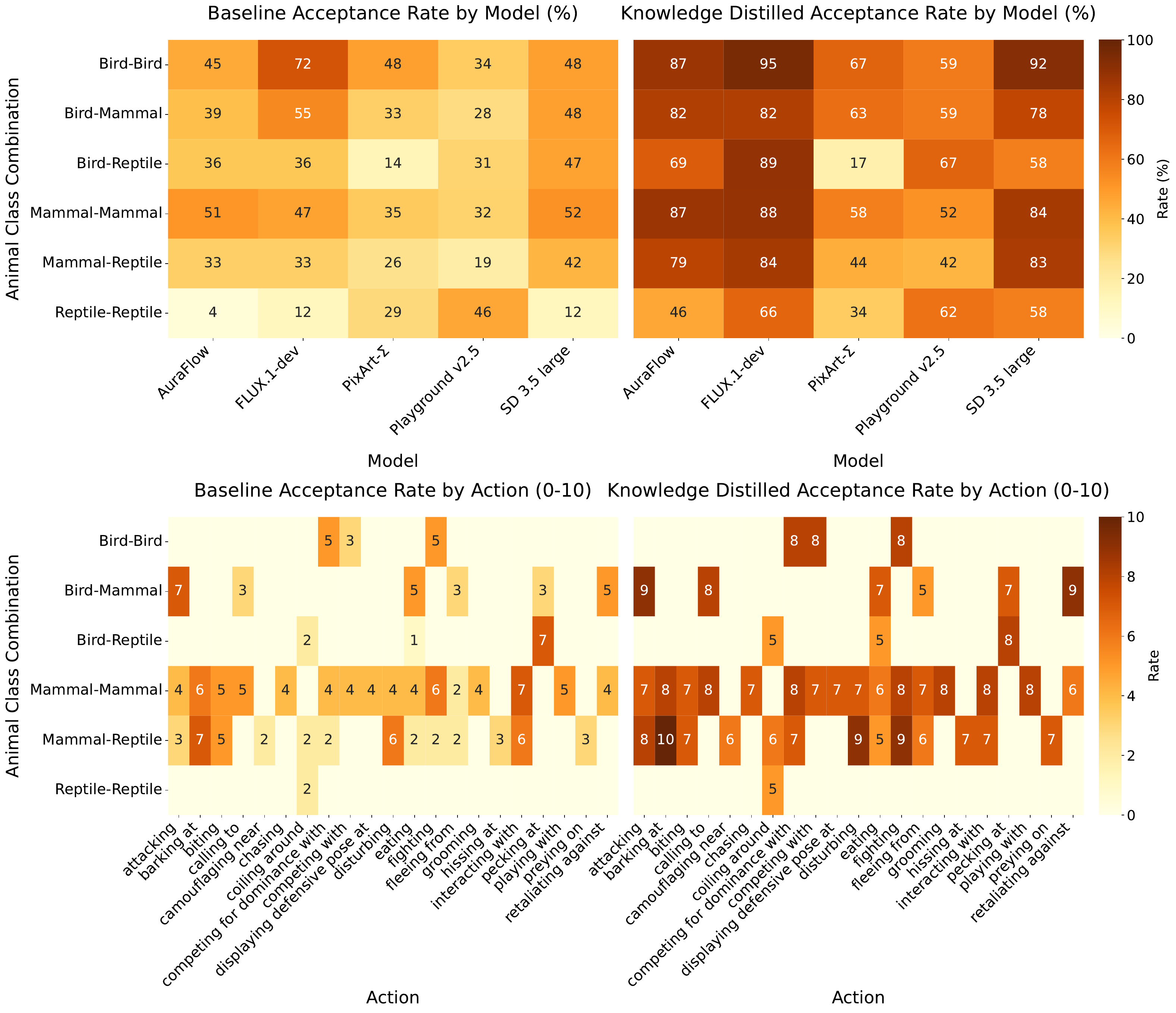}
    \caption{Heatmaps of acceptance rates for baseline and knowledge-distilled prompts. (1a) Baseline acceptance rates by model, (1b) Knowledge-distilled acceptance rates by model, (2a) Baseline acceptance rates by action, and (2b) Knowledge-distilled acceptance rates by action. Comparisons are shown across models, actions, and animal class combinations.}
    \label{fig:accept-rate-detailed}
\end{figure*}

\textbf{Category-Specific Trends}: In Figure \ref{fig:accept-rate-detailed} we elaborate upon the acceptance rate across 2 dimensions,  \textbf{1}. Animal Class vs Model Performance, and \textbf{2}. Animal Class vs Action: Figure \ref{fig:accept-rate-detailed}(1a) breaks down acceptance rates by animal class combinations. We find that birds generally yielded the most accurate depictions, followed by mammals, while models struggled at generating images containing reptiles. FLUX.1-dev excelled in Bird-Bird prompts, reaching a 72\% acceptance rate, and Stable Diffusion 3.5 Large performed best under Mammal-Mammal scenarios (52\%). However, both struggled with reptile-related prompts. We find that Playground v2.5—despite its low overall acceptance—performed comparatively well on Reptile-Reptile prompts (46\%), surpassing even top-performing models in this category. This suggests that some models may have niche strengths or training biases that favor certain animal classes or interactions.

These findings underscore the complexity of action-centric generation tasks. Although certain models achieve moderate success in specific domains (e.g., birds), consistently depicting complex interactions across diverse species remains a significant challenge. 

\subsection{Quantitative Analysis}

Although T2I models have advanced considerably, our evaluations reveal persistent difficulties in accurately depicting complex, action-centric scenes. Figure \ref{fig:accept-rate-detailed}(2a) details acceptance rates across various animal classes and actions, illustrating several recurring issues:

\noindent \textbf{1. Incomplete Depictions:}  We find that a lot of  images lack essential elements of the prompt. For instance, “coiling around” actions often produced headless snakes (Mammal-Reptile acceptance: 17.5\%; Reptile-Reptile: 20.8\%), and “[animal] fleeing from a cobra” frequently omits the cobra entirely. In multiple scenarios, an animal is entirely replaced by another or completely skipped, indicating that models struggle to maintain multiple distinct entities simultaneously.

\noindent \textbf{2. Hybridization of Animals:}  Models occasionally fuse features of different species, yielding unnatural hybrids (e.g., a viper with a duck’s head). For actions such as “pecking at”, "fleeing from", and "calling to" in the Bird-Mammal prompts, the low acceptance rate of 3/10 suggests difficulty differentiating species. Cross-class prompts like “a swan \textit{pecking at} a crocodile” often produces visual blends, undermining species boundaries.

\noindent \textbf{3. Contextual Misrepresentation:} A key challenge lies in accurately rendering the intended relationships and roles specified in a prompt. For example, cross-species interactions frequently exhibit taxonomic bias, with mammalian traits disproportionately emphasized over reptilian ones, regardless of the intended narrative. Consider, for instance, the act of "a Snake coiling around a Gecko". While the physical action of coiling may be represented, images often neglect the gecko's struggle and portrays it as willingly entangled (25\% Mammal-Reptile). It is also overfitted as a negative action instead of neutral. The same taxonomic dominance is given more credence than physical dominance, with small animals able to attack big animals in a spatial area; failing to convey subtle power dynamics (43.8\% Mammal-Mammal; 50.3\% Bird-Bird).

\noindent \textbf{4. Spatial and Positional Inaccuracy:}  Scenarios requiring careful scaling and depth cues are mishandled. “A bird landing on an elephant” often showed disproportionate sizes, while “perched atop a tall giraffe” lacked proper perspective. Such misalignments indicate a struggle to represent realistic spatial relationships.

\noindent \textbf{5. Emotional and Expressive Inaccuracy:}  Prompts implying aggression or social nuance frequently produced incorrect images. For example, “retaliating against” in Bird-Mammal contexts reached only 46.7\% acceptance, rarely capturing the intended hostility. Similarly, “grooming” interactions (35.6\% Mammal-Mammal) lacked the gentle or intimate postures expected.

\noindent \textbf{6. Temporal Dynamics and Action Timing:}  Actions involving movement, such as “chasing,” were often rendered statically. With only 38.7\% acceptance in Mammal-Mammal combinations, dynamic sequences appeared frozen in a single frame, lacking the temporal cues necessary to convey motion and directionality.

Collectively, these challenges underscore that current T2I models struggle with tasks demanding nuanced relational, spatial, emotional, and temporal understanding. Such deficiencies motivate our subsequent efforts to enrich prompts with semantic guidance to improve action depiction.

\section{Improving Action Generation with LLM-guided Knowledge Distillation}
\label{sec:semantic_enrichment}

To address the persistent challenges in T2I generation identified earlier, we propose a training-free prompt enrichment strategy that leverages Large Language Models (LLMs). Specifically, we use GPT-4 \cite{openai2024gpt4technicalreport} to infuse additional semantic guidance into prompts. Rather than opting for retraining T2I models or modifying their architectures, we focus on enhancing the textual inputs directly.\footnote{The total API cost of using GPT-4o for enriching our full 125-prompt dataset (375 calls across three dimensions) was approximately USD~0.20.} This approach is a lightweight and flexible intervention, aiming to provide clearer instructions that help models better capture relational, emotional, and temporal nuances (see Appendix~\ref{subsec:prompting_exp} for few-shot results and Appendix~\ref{subsec:open_vs_closed} for the open- vs.\ closed-source LLM comparison).

\subsection{Dimensions for Prompt Expansion}
We systematically enrich prompts along three key dimensions: \textit{spatial}, \textit{emotional}, and \textit{temporal}. Spatial guidance clarifies relative positioning, size, and depth; emotional cues emphasize behavioral expressions and postures; and temporal hints convey motion and sequential dynamics. By isolating these aspects, we can precisely target common failure modes in T2I generation.

\begin{figure}
    \centering
    \includegraphics[width=\linewidth]{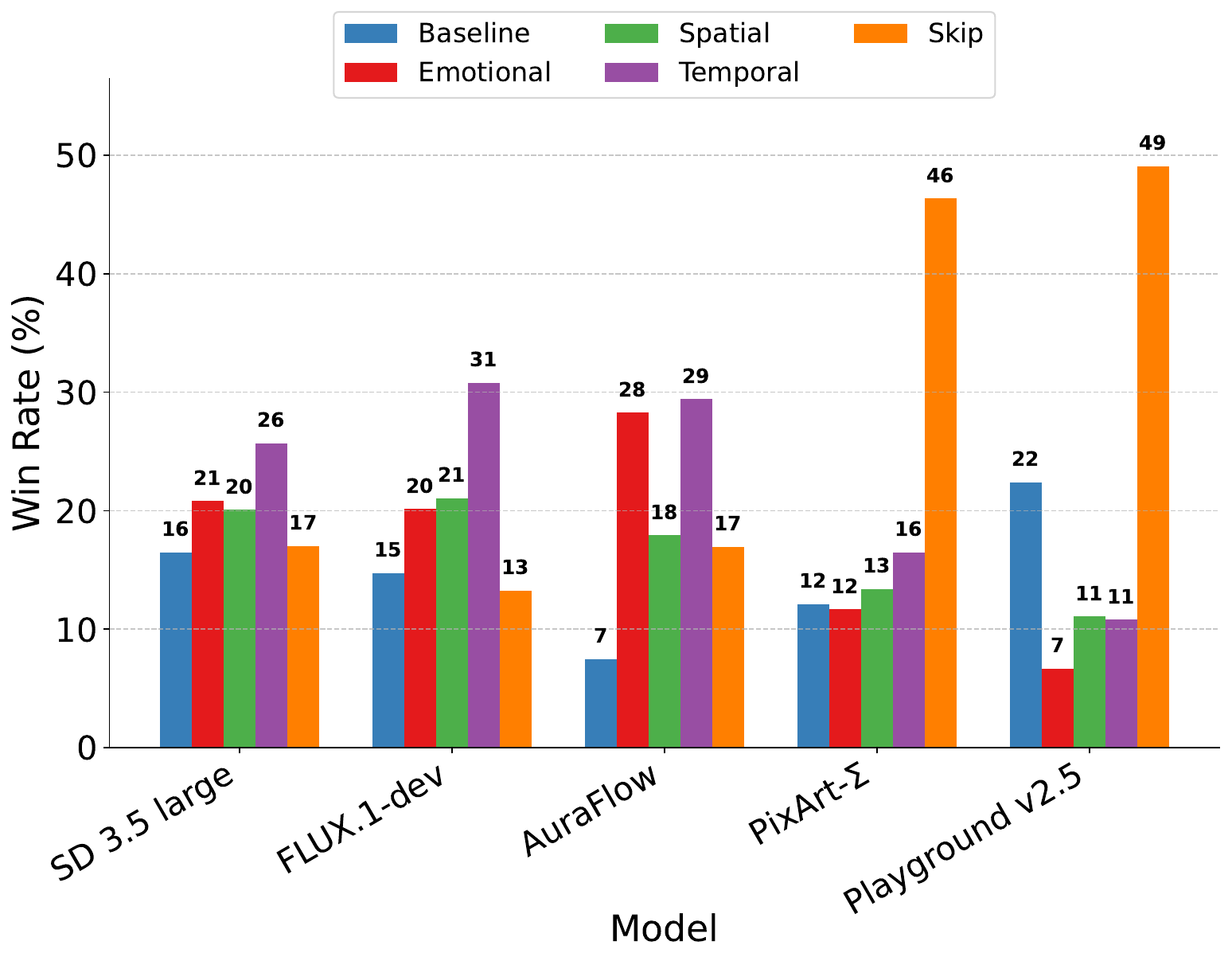}
    \caption{Aggregated user preferences (\%) for image generation models across compositional dimensions. Win rates show how often users preferred a dimension over others. Skip rates represent cases where no dimensions adequately captured action-related elements.}
    \label{fig:win-rate}
\end{figure}
\subsection{Methodology}
For each original prompt, we instruct GPT-4 to add semantic depth tailored to one of the three dimensions. This involves specifying the animals’ relative positions, emotional states, or motion cues more explicitly. The enriched prompts thus serve as more detailed “blueprints” for the T2I model, potentially reducing ambiguity and guiding it toward more accurate renderings.

We provide illustrative templates for each dimension in Appendix \ref{subsec:prompt_guidelines}. These templates demonstrate how a prompt can be transformed to highlight specific spatial, emotional, or temporal aspects without fundamentally altering the underlying scenario. Importantly, these transformations are prompt-specific: each prompt’s enrichment depends on its initial wording and context. By applying the template guidelines flexibly, we can adapt the semantic enrichment process to a wide range of action-centric scenes, ensuring that the resulting prompts remain coherent, contextually relevant, and aligned with the desired narrative.

\begin{figure*}[h]
    \centering
    \includegraphics[width=\linewidth]{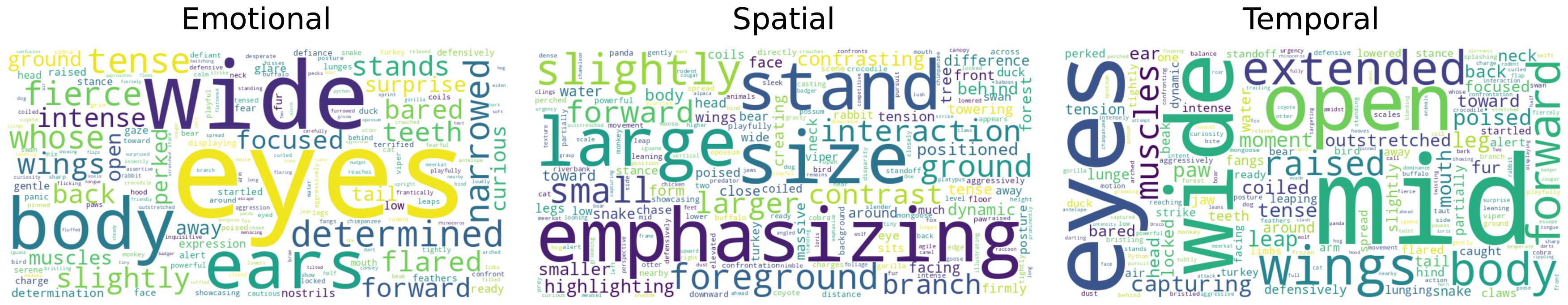}
    \caption{Word clouds summarizing key semantic elements enriched in prompts for each dimension: Emotional (left), Spatial (center), and Temporal (right).}
    \label{fig:wordclouds}
\end{figure*}

\subsection{Evaluation of Distillation Techniques}

Knowledge distillation significantly enhances the performance of T2I models, particularly in capturing temporal, emotional, and spatial nuances. Figure \ref{fig:win-rate} illustrates the win rate of each model across various dimensions. Stable Diffusion 3.5 Large, Flux.1-dev, and AuraFlow demonstrate notable user preference for LLM-guided enriched prompts. Temporal Distillation emerges as the most preferred option across all models, followed by Emotional and Spatial Distillation. Conversely, PixArt-$\Sigma$ and Playground v2.5 exhibit limited efficacy in utilizing descriptive prompts. Subsequent paragraphs provide an in-depth analysis of each dimension's performance across different animal class pairs and actions. Our word cloud analysis (Figure \ref{fig:wordclouds}) reveals the most frequent terms in our enriched prompt database across dimensions. Emotionally Distilled prompts feature a high frequency of expressive terms, effectively capturing subject emotions and moods. Spatially Distilled prompts emphasize precise locational descriptors, ensuring accurate spatial relationships. Temporal Distillation incorporates temporal markers, enhancing the depiction of dynamic sequences. Additionally, we conduct a POS Analysis, detailed in Appendix \ref{subsec:prompt_pos_analysis}.

\noindent\textbf{Temporal Distillation:}  
We find that temporal enrichment causes the largest improvement in performance. For example, competitive actions, which often involve nuanced sequences of dominance and retaliation, see a +274\% improvement with temporal prompts. Bird-Bird interactions, known for intricate hierarchical displays, achieve a +365\% gain. Specific actions like “hissing at” and “retaliating against” improve by +431\% and +363\%, respectively, highlighting the critical role of action timing and motion cues in disambiguating complex behaviors.

\noindent\textbf{Emotional Distillation:}  
Emotional guidance refines subtle behavioral and expressive details, substantially boosting fidelity in close-range or tension-filled scenarios. Feeding actions, which depend on accurately depicting predatory and defensive postures, benefit by +223\%, while “chasing” and “coiling around” improve by +397\% and +382\%, respectively. In reptile-reptile interactions, emotional cues lead to a +891\% improvement, underscoring how clear affective states help models represent inherently less familiar or visually subtle animal dynamics.

\noindent\textbf{Spatial Distillation:}  
Spatial enrichment ensures correct positional relationships and size contrasts. While its impact is modest in dynamic scenarios, it still provides meaningful gains for actions reliant on correct vantage points. For example, “calling to” improves by +307\%, and “hissing at” sees a +292\% gain under spatial prompts. These enhancements confirm that clearly specifying spatial arrangements can complement temporal and emotional cues, particularly for stationary or less overtly dynamic interactions.

\noindent\textbf{Baseline and Category Dependencies:}  
Interestingly, certain categories remain challenging, with social actions showing relatively modest gains (+37\%), and baseline prompts outperforming enriched ones in some aggressive and social scenarios. Bird-Reptile interactions stand out at baseline (0.278), indicating that even without enrichment, some combinations are inherently easier.

Overall, these quantitative insights validate the qualitative claims. Temporal cues best tackle dynamic and abstract actions, emotional details help articulate close-range or expressive interactions, and spatial guidance refines positional accuracy. While no single technique solves all shortcomings, dimension-specific enrichment—particularly temporal—offers a significant step toward more nuanced, contextually accurate T2I image generation. 

\subsection{What about automated metrics?}

\begin{figure*}
    \centering
  \begin{subfigure}{0.5\textwidth}
    \includegraphics[width=\linewidth]{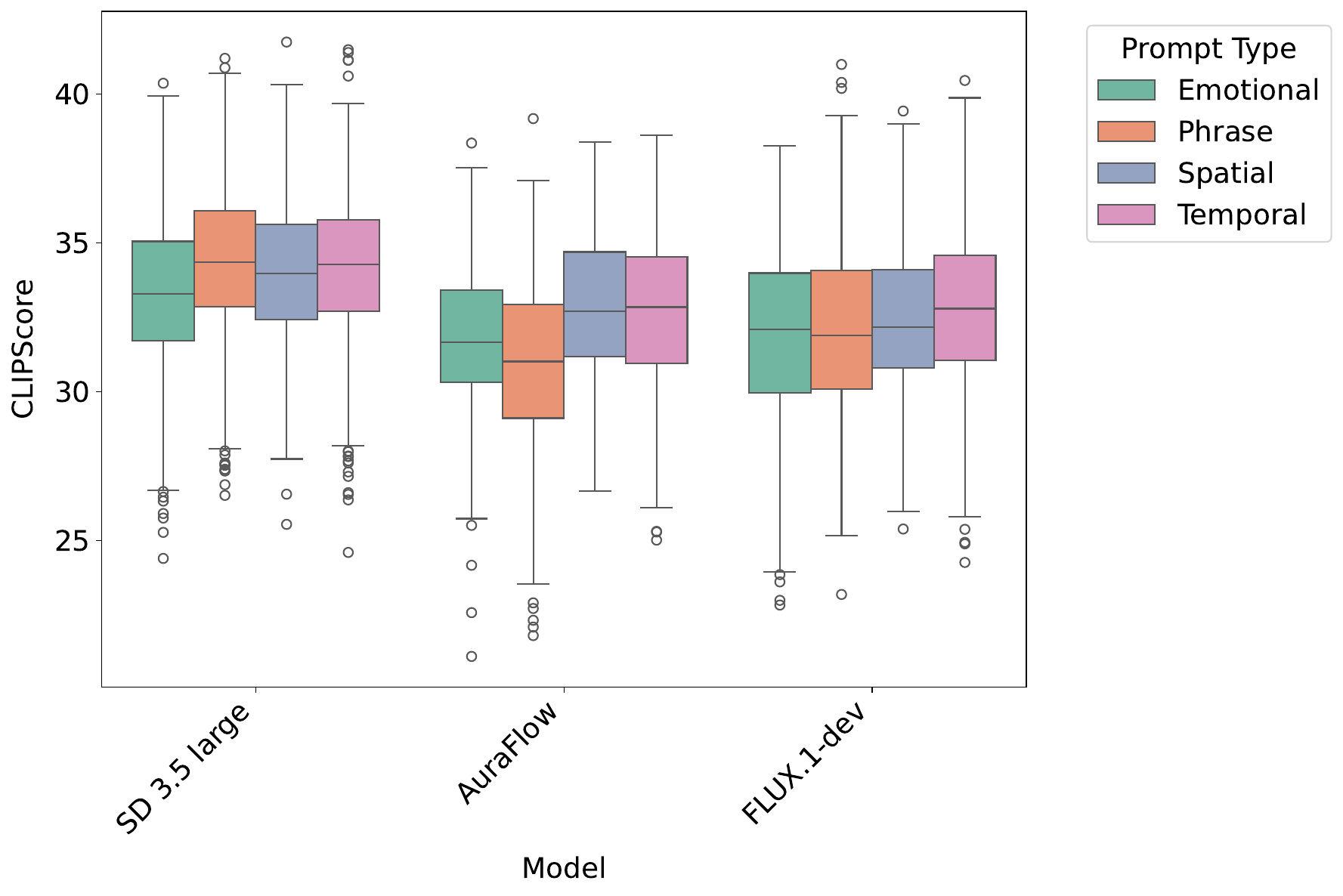}
    \caption{CLIPScore (higher is better)}
    \label{fig:sub1}
  \end{subfigure}%
  \hfill
  \begin{subfigure}{0.5\textwidth}
    \includegraphics[width=\linewidth]{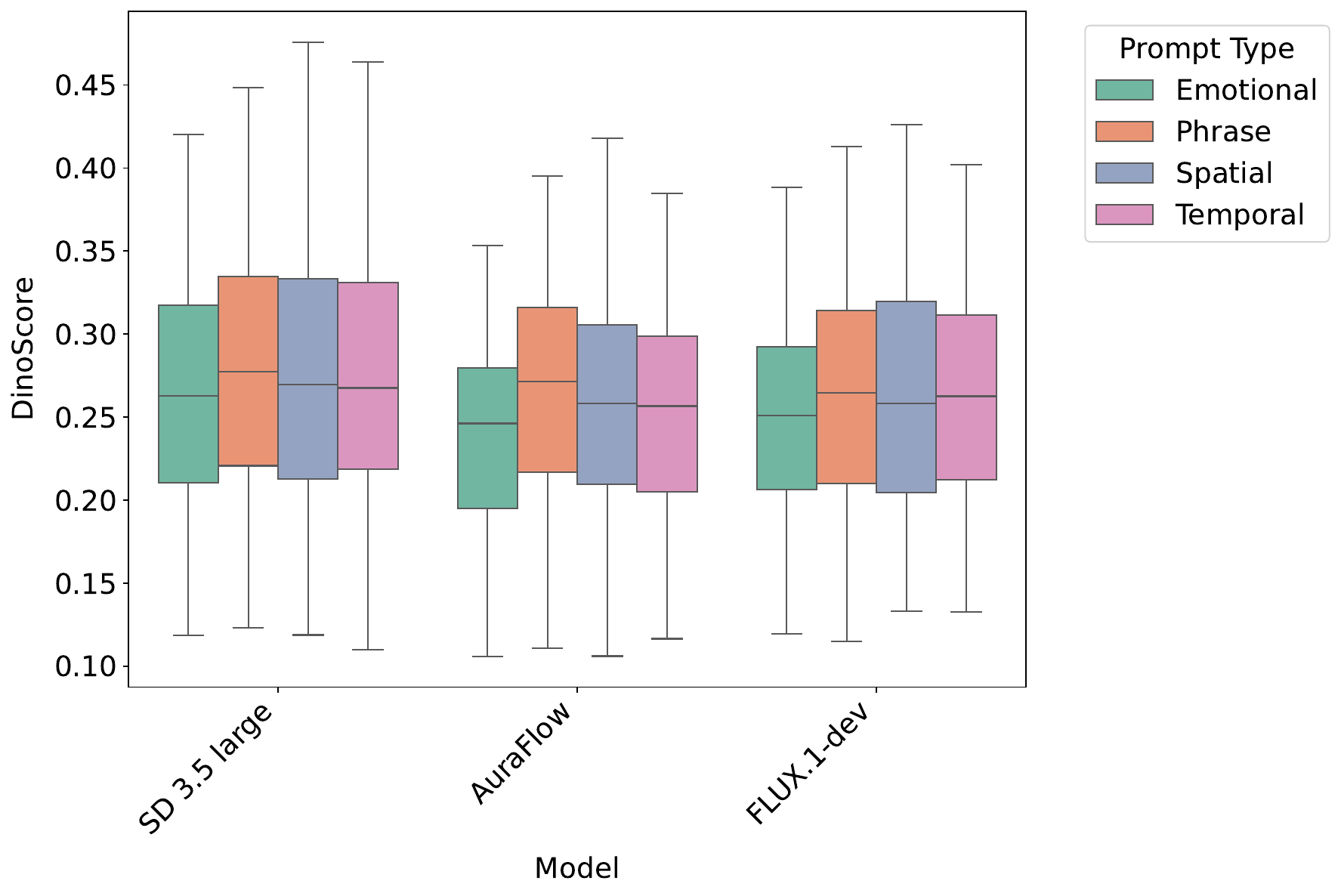}
    \caption{DinoScore (higher is better)}
    \label{fig:sub2}
  \end{subfigure}
  \caption{Performance of Automated Evaluation Metrics on the AcT2I Benchmark. Neither metric aligns well with human annotator preferences.}
  \label{fig:automated_mertics}
\end{figure*}

We explore CLIPScore \cite{hessel2021clipscore} and a DINOv2\cite{oquab2023dinov2} based metric (DinoScore) to automatically evaluate the generated images. The objective was to assess whether automated metrics are reliable for evaluating action generation in images. CLIPScore is a reference-free evaluation metric that leverages the capabilities of the CLIP model to assess semantic alignment between textual descriptions and images. The DINOv2-based pipeline has 2 steps : \textbf{1} Extracting the most relevant frame from Animal Kingdom videos for a given action (using CLIPScore), and \textbf{2} comparing the DINOv2 features of the extracted frame and the generated image of the T2I models. As shown in Figure~\ref{fig:automated_mertics}, automated metrics exhibit minimal differentiation between dimensions and fail to correlate with human evaluations on the AcT2I benchmark. Although the means in Figure~\ref{fig:automated_mertics} appear almost identical, this reflects a shared insensitivity to fine-grained action cues rather than genuine agreement; a targeted probe (Appendix~\ref{sec:metric-sanity-example}) shows that both metrics assign nearly the same scores to correct and mismatched captions. This discrepancy underscores the need for more sophisticated metrics that better align with human perceptual judgment. We discuss this in more depth in Appendix~\ref{sec:automated-eval}.

\paragraph{Note on multimodal LLM evaluators.}  
A pilot study with the VQA-capable model \texttt{llava-v1.6-vicuna-13b-hf} is discussed in Appendix \ref{sec:mm-eval}.  Its limited accuracy ($\approx$70 \% overall, 62 \% on semantic queries) suggests current multimodal LLMs still struggle with interpreting fine-grained action semantics.

%% file: tables/prompt_examples.tex
\begin{table}[t]
\centering
\resizebox{\linewidth}{!}{
\begin{tabular}{@{}llll@{}}
\toprule
A &  B  &  Action  &  Text \\
\midrule
Goose & Turkey & competing for dominance with & A goose competing for dominance with a turkey \\
Boar & Giraffe & retaliating against & A boar retaliating against a giraffe \\
Weasel & Snake & fleeing from & A weasel fleeing from a snake \\
Duck & Duck & fighting & A duck fighting a duck \\ 
Gorilla & Dog & grooming & A gorilla grooming a duck \\ 
\bottomrule
\end{tabular}
}
\caption{
Examples of text inputs from the AcT2I dataset for a pair of animals (A, B) and an action between them.
}
\label{tab:prompt_examples}
\end{table}

%% file: sec/4_conclusion.tex
\section{Conclusion}
\label{sec:conclusion}

Despite significant advancements in Text-to-Image (T2I) synthesis, current models exhibit limitations in accurately representing nuanced actions, highlighting a gap between model capabilities and real-world expectations. Our evaluation reveals that state-of-the-art models achieve only a 48\% acceptance rate, underscoring the difficulty in capturing the implicit visual cues crucial for representing dynamic scenarios.

To address this challenge, we introduced knowledge-distilled techniques targeting three key dimensions: temporal dynamics, emotional expressiveness, and spatial relationships. Temporal distillation emerged as the most impactful, significantly enhancing the depiction of dynamic actions. Emotional and spatial distillations complemented this by refining subtle behavioral and positional elements, respectively.

While automated metrics like CLIPScore and DinoScore offer valuable insights, they fall short in capturing the nuanced improvements achieved through our techniques. Human evaluations remain the gold standard for assessing semantic fidelity and realism in complex T2I outputs.

Future research should focus on enhancing relational and temporal grounding in vision-language (VL) models to better capture the implicit visual cues critical for nuanced action representation. Furthermore, the development of robust automated metrics capable of accurately evaluating complex T2I outputs remains a crucial area for future exploration, ensuring that progress in this field can be effectively measured and validated.

%% file: sec/5_limitations.tex
\section{Limitations}

While this study demonstrates the efficacy of training-free semantic enrichment, two primary limitations merit consideration. First, the observed performance gains are contingent upon the underlying Text-to-Image (T2I) model's capacity to effectively leverage densely enriched prompts. The extent of this capacity may vary depending on the initial prompt complexity and the quality of the Large Language Model (LLM)-generated output. Second, the evaluation methodology relies heavily on human assessment due to the inherent limitations of current automated metrics, which are often inadequate for capturing subtle semantic nuances comprehensively. While our findings indicate significant improvements in contextual understanding for T2I models, they also raise potential societal concerns, including the potential misuse of more realistic imagery and the propagation of inherent biases present within the training data. Future research must address these ethical considerations to ensure responsible applications of these techniques and mitigate potential negative consequences.

\section*{Acknowledgment}
This work was supported by NSF RI grant \#2132724. 
We thank the NSF NAIRR initiative, Research Computing (RC) at Arizona State University (ASU) for their generous support in providing computing resources. 
The views and opinions of the authors expressed herein do not necessarily state or reflect those of the funding agencies and employers.

%% file: sec/z_appendix.tex
\appendix
\section*{Appendix}
\FloatBarrier

\section{Dataset Coverage and Axis-wise Performance}
\label{sec:dataset_coverage}
\subsection{Prompt taxonomy and distributions}

We annotate all 125 prompts along four orthogonal axes. Table~\ref{tab:axis-rules} defines the axes with category rules and examples; Table~\ref{tab:axis-dist} summarizes coverage and design rationale.

\input{tables/axis_rules}
\input{tables/axis_dist}

\subsubsection{How the axis values were computed.}
\begin{enumerate}\itemsep2pt
\item Each prompt’s \emph{verb} determines its Emotional, Spatial, and Temporal labels via the mapping above.
\item Interaction rarity is assigned manually using the provided guideline (routine / occasional / improbable).
\item Percentages in Table~\ref{tab:axis-dist} are simple counts over the 125 prompts.
\end{enumerate}

\input{tables/axis_acc}

\subsection{Notes on interpretation (with entropy)}
We do not observe a strong monotonic trend by interaction rarity (frequent/rare/very-rare are similar). Slight variations appear by intent and topology: defensive scenes underperform communicative/affiliative, and pursuit/avoidance (dynamic) lags static distal or proximal cases. Overall, action depiction challenges persist across bins rather than being confined to a narrow slice; enrichment improves all bins. Shannon entropy for the primary-verb distribution is $0.97$, and for the other axes it is $\geq 0.82$, indicating that no axis is dominated by a single category and confirming balanced coverage that reduces the risk of any single bin driving these effects.
\section{Addressing Generalization Limitations}
\label{sec:generalization}
\begin{figure*}[h]
  \centering
  \begin{subfigure}[t]{0.45\textwidth}
    \includegraphics[width=\textwidth]{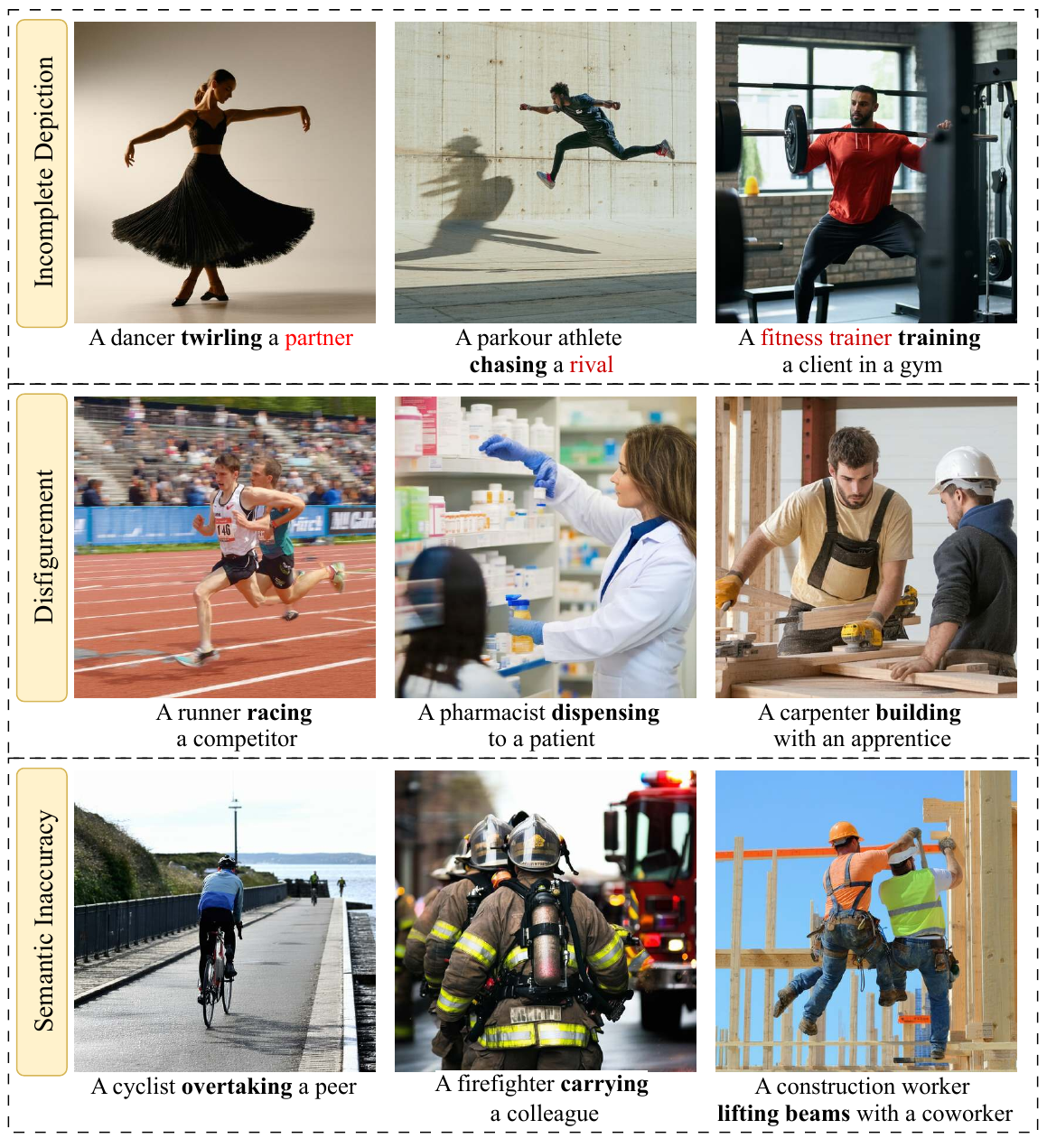}
    \caption{Examples of generation failures for human actions.}
    \label{fig:human_samples_a}
  \end{subfigure}
  \hfill
  \begin{subfigure}[t]{0.54\textwidth}
    \includegraphics[width=\textwidth]{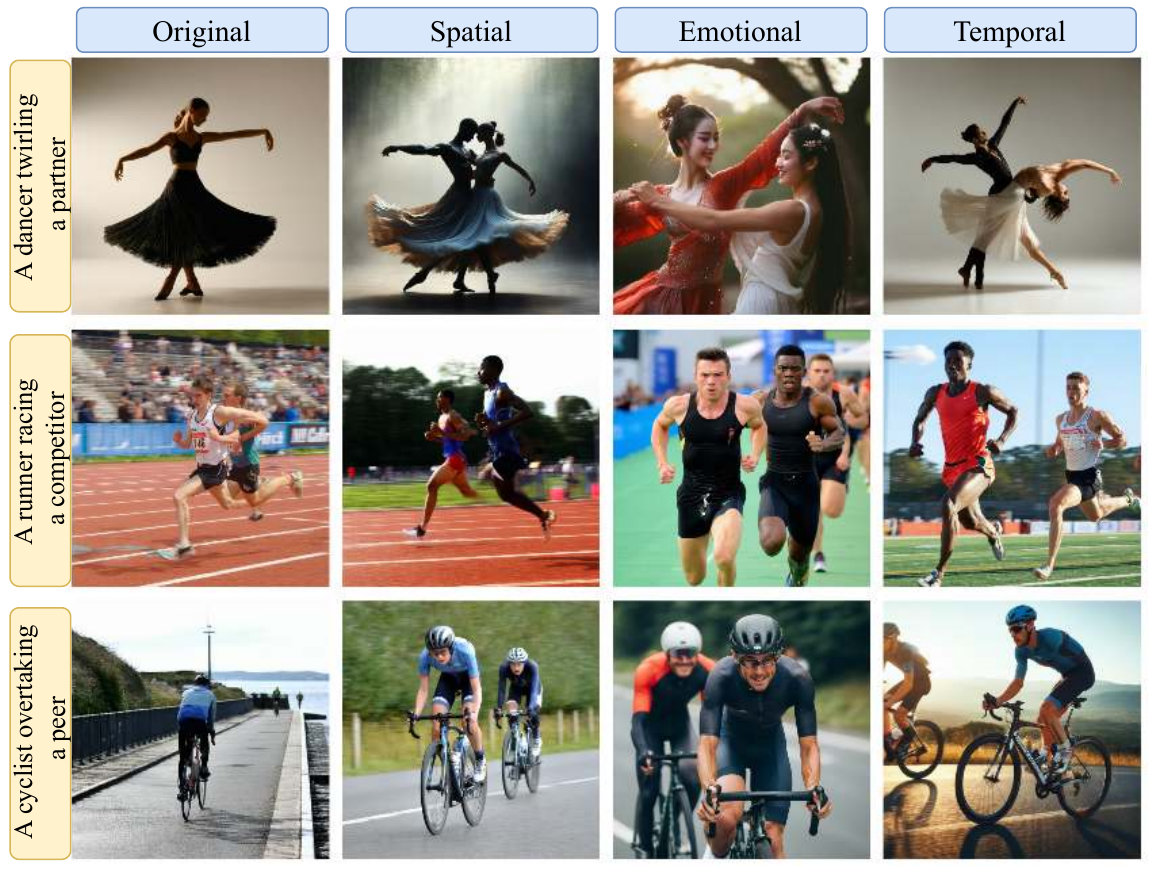}
    \caption{Semantic improvements achieved through knowledge distillation.}
    \label{fig:human_samples_b}
  \end{subfigure}

  \caption{Challenges in generating images of human actions with Stable Diffusion v3.5 Large. Panel (A) highlights common errors— incomplete depictions (missing subjects or objects), disfigurement (physical anomalies), and semantic inaccuracies (misrepresented actions). Despite these errors, panel (B) demonstrates the generalization capabilities of our technique on human samples.}
  \label{fig:human_sample}
\end{figure*}

To evaluate action depiction, we aimed to assess T2I models' ability to generate images where all subjects were accurately represented. Additionally, our choice of subjects prioritized those with high affordance, capable of performing a diverse set of actions and assuming unique roles with distinct mappings. However, the current capabilities of T2I models, when tasked to generate all subjects correctly, introduce more issues. We found that generating images of humans remains a challenge, with common errors including incomplete depictions and subject disfigurement. These errors directly impact evaluation because we can only thoroughly critique semantic inaccuracies after the subjects themselves are rendered correctly. We share examples of such errors in Action depiction with two Human subjects using Stable Diffusion 3.5 Large in Figure \ref{fig:human_samples_a}. 

To minimally probe generalization beyond animals, we tested 10 human--human relations with Stable Diffusion~3.5 using baseline vs.\ enriched prompts; representative items include: dancer--partner (twirling), runner--competitor (racing), firefighter--colleague (carrying), boxer--rival (sparring), climber--teammate (belaying), cyclist--peer (overtaking), chef--sous-chef (tossing ingredients), construction worker--coworker (lifting beams), parkour athlete--rival (chasing), and kayaker--teammate (paddling alongside). Across these examples, enrichment 9 out of 10 times improved temporal/spatial/emotional fidelity; qualitative illustrations appear in Fig.~\ref{fig:human_samples_b}.

Likewise, prompts involving abstract concepts are problematic because their many visual interpretations make objective evaluation much harder, leading us to focus on more grounded scenarios. Due to their inherent characteristics, animals provide an ideal test case for evaluating the generalization performance of T2I models, making them our primary choice for this study.

\section{Qualitative Evidence of Semantic Enrichment}
\label{sec:qualitative-grid}

\begin{figure*}
    
    \centering
    \includegraphics[width=\textwidth]{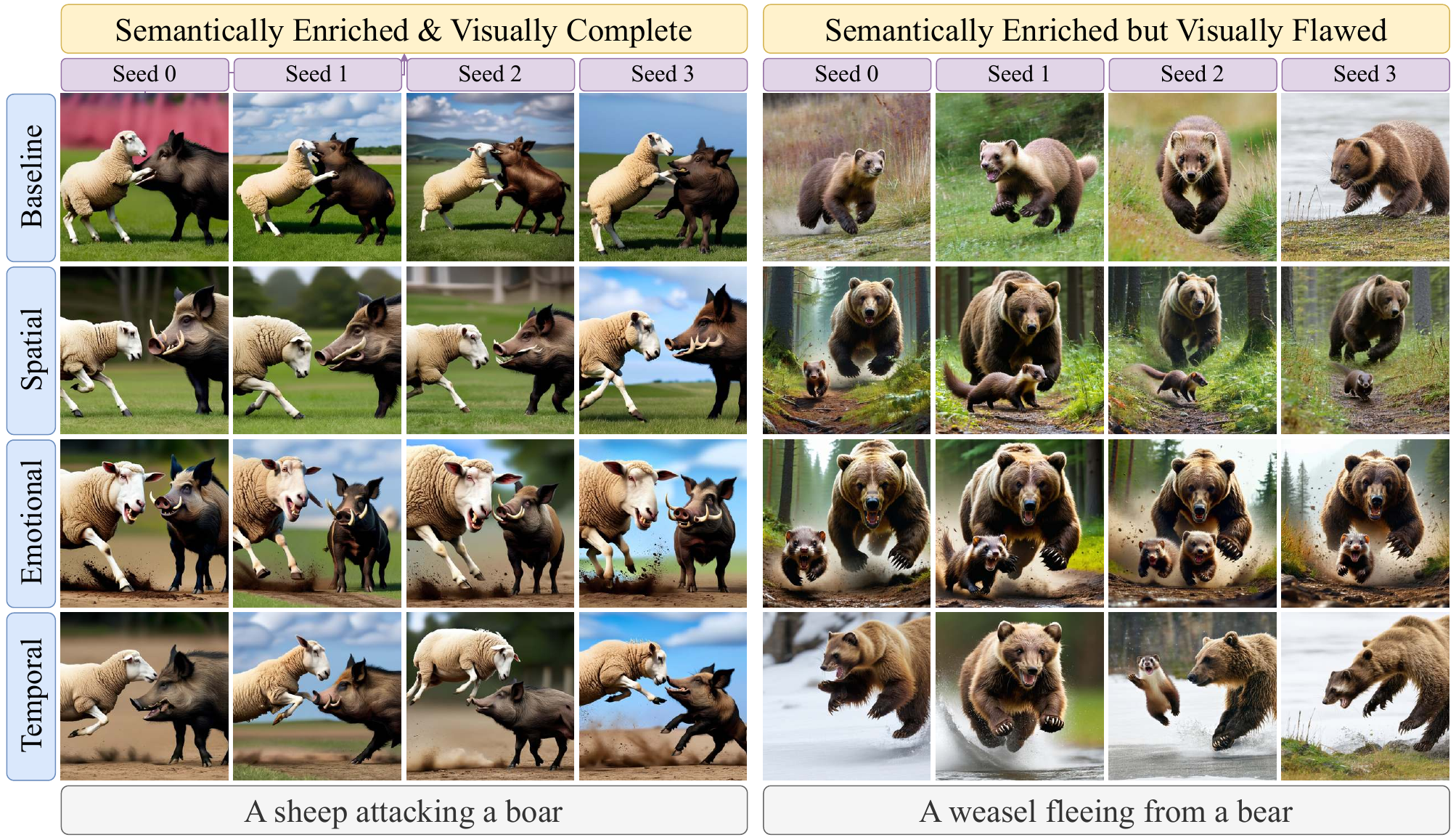}
    \caption{Qualitative grid illustrating our LLM-guided prompt enrichment.  
\textbf{Left block (“Semantically Enriched \& Visually Complete”)}: spatial layout, emotional cues, and temporal framing all align with the intended action.  
\textbf{Right block (“Semantically Enriched but Visually Flawed”)}: the same semantic cues are present, but pixel-level errors such as missing subjects or distortions remain, highlighting that enrichment injects reliable semantics even when rendering fidelity fails.}
\label{fig:grid_semantics}
\end{figure*}

Figure \ref{fig:grid_semantics} presents eight representative generations produced with our LLM-guided prompt enrichment.  The images are grouped into two categories:

\vspace{0.3em}
\noindent\textbf{Semantically Enriched \& Visually Complete}  
(left block, four examples)\,—
spatial layouts, emotional expressions, and temporal framing all align with the target action, yielding high-fidelity results (e.g., \emph{“a sheep attacking a boar”}).  

\vspace{0.3em}
\noindent\textbf{Semantically Enriched but Visually Flawed}  
(right block, four examples)\,—
the same semantic cues are present, yet the base T2I model introduces pixel-level errors such as missing subjects or anatomical distortions (e.g., \emph{“a weasel fleeing from a bear”}).  

Across both groups, three high-level dimensions are consistently evident:

\begin{itemize}
\item \textbf{Spatial relationships}: correct relative placement and orientation of agents.
\item \textbf{Emotional cues}: facial expressions or body posture that match the action context.
\item \textbf{Temporal framing}: a frame that captures the peak moment of the action.
\end{itemize}

The persistence of these cues in visually flawed outputs indicates that semantic enrichment operates independently of pixel-level rendering quality, complementing the quantitative gains reported in Section \ref {sec:semantic_enrichment}.

\section{More results on Automated Metric}
\subsection{DinoScore Evaluation}
\label{sec:automated-eval}
We collect human annotation preferences and derive a consensus from the three reviews for each sample. The preferred dimension is then compared to the highest CLIPScore and DinoScore, respectively. Table \ref{tab:dino_clip_eval} presents the alignment of these two automated evaluation metrics with human preferences. Given four possible choices, the baseline alignment is 0.25. The highest alignment observed among the top three models is 0.38 for CLIPScore and 0.30 for DinoScore. These results indicate that while there is some overlap, the metrics exhibit significant limitations in capturing subtle semantic improvements.
\input{tables/dino_clip_eval}

\subsection{Action-specific probe for metric granularity}
\label{sec:metric-sanity-example}

We generated an image with Stable Diffusion~3.5 for the prompt \emph{“a cat chasing a mouse”}. Table \ref{tab:probe_scores} lists CLIPScore and DinoScore for three candidate captions— \textit{chasing} (correct), \textit{observing from afar}, and \textit{attacking}.
The correct caption scores highest in both metrics, yet the margin over incorrect captions is small (< 5 \% for CLIPScore, < 0.04 absolute for DinoScore), confirming that the metrics capture high-level entity alignment but struggle with nuanced action semantics.

\begin{figure}[t]
  \centering
  \includegraphics[width=0.65\columnwidth]{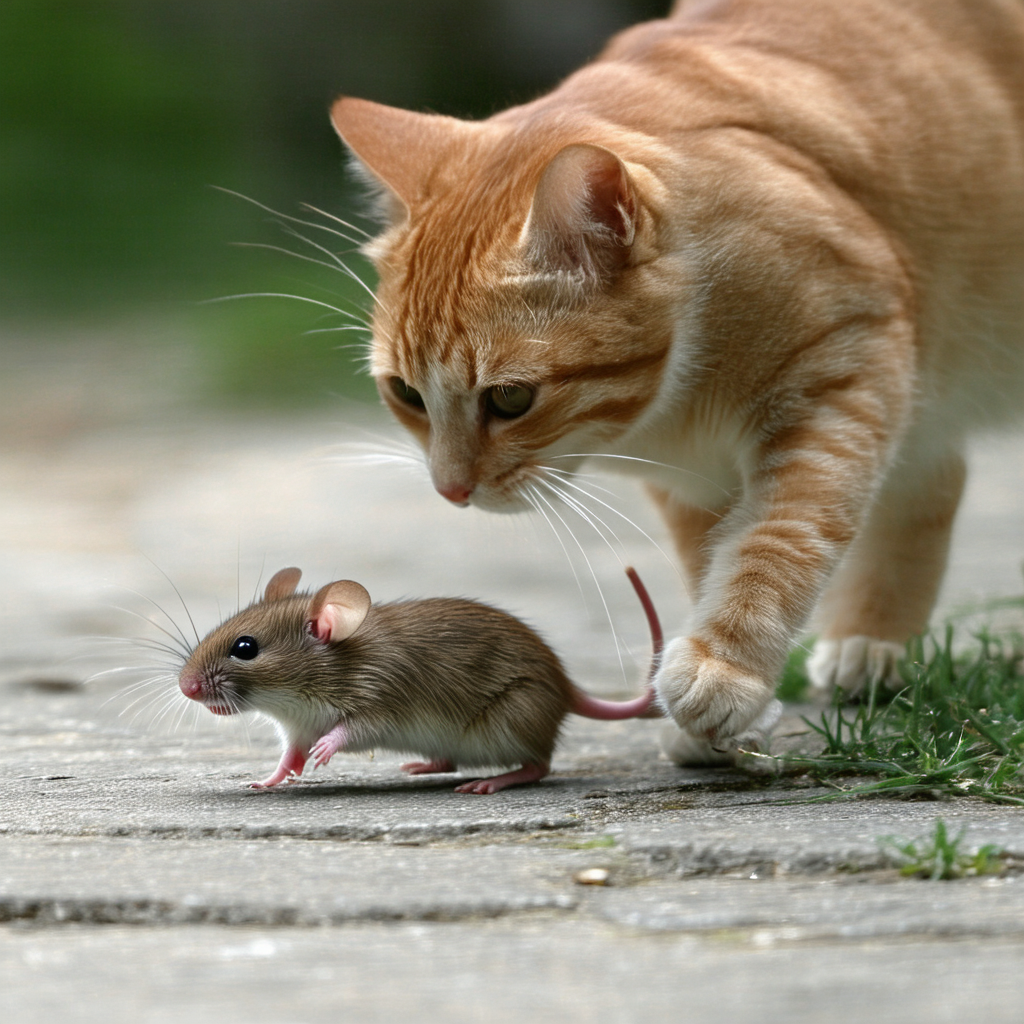}
  \caption{Stable Diffusion 3.5 image generated for “cat chasing a mouse.”}
  \label{fig:probe_img}
\end{figure}

\begin{table}[t]
  \centering
  \small
  \begin{tabular}{@{}p{0.55\linewidth}cc@{}}
    \toprule
    \textbf{Candidate caption} & \textbf{CLIP}$\uparrow$ & \textbf{DINO}$\uparrow$\\
    \midrule
    Cat \textbf{chasing} a mouse   & 26.9 & 0.257\\
    Cat \textbf{observing} from afar & 25.9 & 0.286\\
    Cat \textbf{attacking} a mouse  & 25.6 & 0.290\\
    \bottomrule
  \end{tabular}
  \caption{CLIPScore and DinoScore for the probe image (Figure \ref{fig:probe_img}).  The small gaps between correct and mismatched captions illustrate each metric’s coarse granularity.}
  \label{tab:probe_scores}
\end{table}

\subsection{Multimodal LLM Evaluation}
\label{sec:mm-eval}
We also explored using the multimodal LLM \texttt{llava-v1.6-vicuna-13b-hf} as a VQA-style evaluator.  For each prompt, we automatically generated ten tailored questions covering salient scene attributes.  On a proof-of-concept (POC) set of eight prompts, LLaVA’s answers matched human annotations 70\% of the time; accuracy dropped to 62\% on questions probing nuanced semantic details (e.g.\ object–action relations).  Given this modest performance and the cost of querying LLMs, we did not pursue this avenue further at scale.

\section{Prompt based Analysis}
\subsection{Action Template Taxonomy}
\label{subsec:action_templates}

In Table~\ref{tab:action_templates}, we share the action templates used to generate prompts, categorized by their plausibility tiers (Highly Plausible, Moderately Plausible, and Less Plausible). These templates guided the selection of animals and actions to ensure a broad range of complexity and contextual requirements.

\input{tables/action_template}

\subsection{Prompt Distillation Guidelines}
\label{subsec:prompt_guidelines}
Below are the guidelines we used to enrich prompts with spatial, temporal, and emotional details with an average token count of 47, 42, and 46, respectively. These were applied using an LLM (GPT-4o) to create enriched versions of the original prompts, providing more explicit cues that aid T2I models in generating contextually accurate images. Through prompts, LLM was instructed to enhance prompts for text-to-image tasks through knowledge distillation in [dimension], followed by an explanation of what was expected. Each instruction concluded with a message of keeping the enhanced prompt concise yet detailed and aiming for approximately 50-70 tokens while prioritizing clarity over length.

\begin{enumerate}
    \item Spatial Relationships and Composition:
    \begin{tcolorbox}[boxsep=0pt,boxrule=1pt]
        \scriptsize
        \texttt{Make implicit spatial details explicit to improve the prompt, while keeping it concise and focused. Pay attention to:\\
        \textbf{Positional Accuracy}: Clearly specify how animals are positioned relative to each other.\\
        \textbf{Depth and Perspective}: Indicate scaling and perspective for appropriate distance and interaction.\\ 
        \textbf{Example}: Instead of "a bird lands on an elephant", say "a small bird gently lands atop a towering elephant's back, highlighting their size difference".}
    \end{tcolorbox}

    \item Temporal Dynamics and Action Timing:
    \begin{tcolorbox}[boxsep=0pt,boxrule=1pt]
        \scriptsize
        \texttt{Make implicit temporal and action details explicit to improve the prompt, while keeping it concise and focused. Emphasize:\\
        \textbf{Optimal Freeze-Frame Selection}: Capture the most expressive moment of the action.\\
        \textbf{Motion Representation}: Use visual cues like dynamic posture to imply movement.\\
        \textbf{Example}: Instead of "a cheetah chases a gazelle", say "a cheetah mid-stride with muscles tensed, closely pursuing a gazelle in full sprint".}
    \end{tcolorbox}

    \item Emotional and Expressive Details:
    \begin{tcolorbox}[boxsep=0pt,boxrule=1pt]
        \scriptsize
        \texttt{Make implicit emotional details explicit to improve the prompt, while keeping it concise and focused. Include:\\
        \textbf{Facial Expressions}: Depict emotions appropriate to the action.\\
        \textbf{Body Language}: Use posture and movement to enhance emotional portrayal.\\
        \textbf{Example}: Instead of "a puppy chases a kitten", say "a playful puppy with a wagging tail chases a kitten that's glancing back with a mischievous grin".}
    \end{tcolorbox}
\end{enumerate}

\subsection{POS Tag Analysis}
\label{subsec:prompt_pos_analysis}
Through POS tagging, we analyzed the prompts to interpret the prompt distillation. We discovered that all the enriched prompts reduced the usage of proper nouns by 70\%-80\% while verbs and nouns increased by 5x-6x and 10x-15x, respectively, across all dimensions. Intra-dimensional analysis revealed that adjectives were 1.5x-2x more frequent in the Emotional dimension compared to other dimensions, while the Spatial dimension exhibited significantly higher usage of determiners, adposition, adverbs, and pronoun-particles. Overall, the top 10 most frequently used adjectives and verbs across each dimension were found to be in alignment with the intended meaning of each.

\subsection{Few Shot Experiments}
\label{subsec:prompting_exp}
We conducted a preliminary experiment employing the few-shot prompting technique using GPT 4o and Gemini 2.0 Flash, utilizing three instances of original enriched prompts. The results of a blind review comparison between images generated from original enriched prompts and few-shot outputs of both models indicated comparable performance.

\subsection{Open Source vs Closed Source LLMs}
\label{subsec:open_vs_closed}
The cost of LLM APIs remains a key concern for the practical utility of our technique. To address this, we conducted a small‑scale analysis comparing the closed‑source model GPT‑4o with the open‑source \texttt{meta‑llama/Llama‑3.3‑70B‑Instruct}. A blind review reveals that both models perform comparably, thereby alleviating cost‑related concerns.
\section{Open Source vs.\ Closed Source LLMs}
\label{subsec:open_vs_closed}

\paragraph{Goal.}
To assess whether an open-source LLM can replace a closed-source model for prompt enrichment without materially affecting outcomes or cost.

\paragraph{Setup.}
We compare \texttt{GPT-4o} (closed) and \texttt{meta‑llama/Llama‑3.3‑70B‑Instruct} (open), applying the same enrichment instructions from Appendix ~\ref{subsec:prompt_guidelines} (Spatial/Emotional/Temporal). For each base prompt, both LLMs produce a single enriched prompt.

\paragraph{T2I Backbone \& Generation.} 
Stable Diffusion 3.5 Large; 1 image per (prompt, LLM) pair; fixed random seed per prompt.

\begin{figure*}[t]
  \centering
  \includegraphics[width=\linewidth]{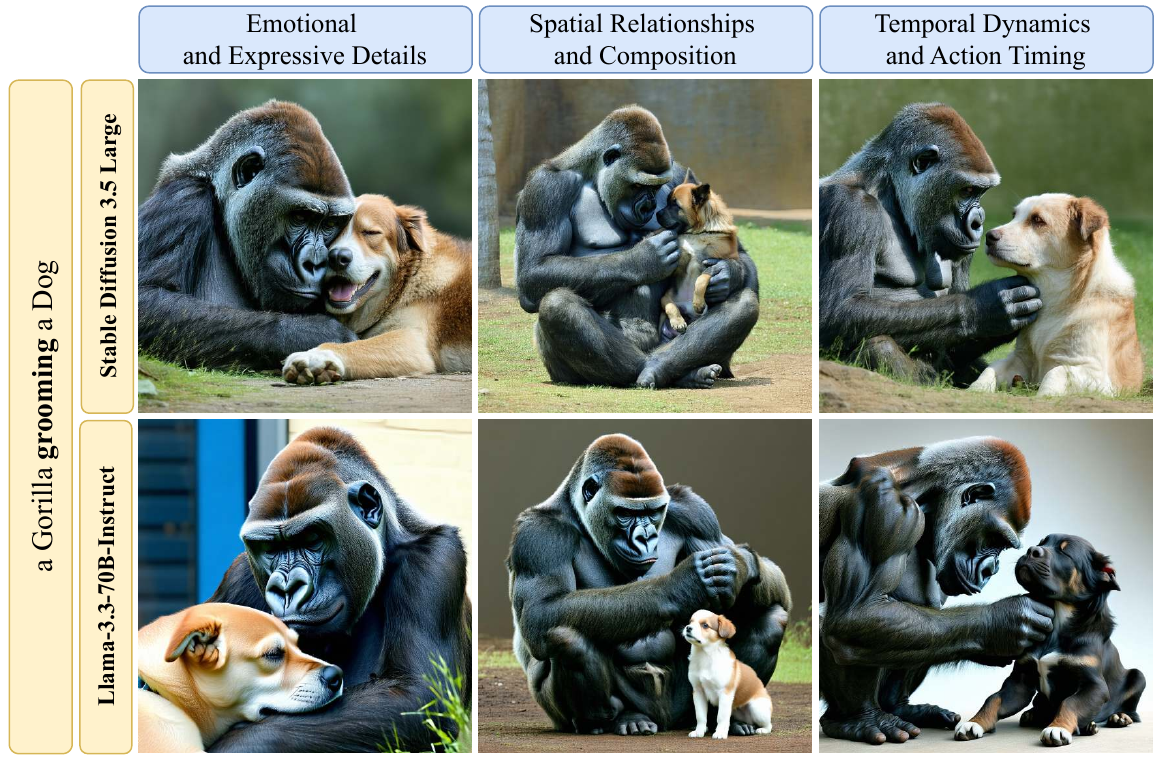}
  \caption{Qualitative example (``\emph{a Gorilla grooming a Dog}''): both variants convey the target relation; differences are subtle.}
  \label{fig:llama}
\end{figure*}

\paragraph{Evaluation (pairwise human preference).}
In a pairwise human–preference study over \textbf{20} prompts\footnote{This 20-prompt subset represents \textbf{17/25} action types and was hand-picked to maintain rough diversity across the four axes: interaction rarity, emotional valence, spatial topology, and temporal extent.}, three independent annotators (same pool as the main study) selected which image better depicted the described action; we aggregate by majority vote per prompt (no ties occurred). Images generated from \texttt{GPT-4o}-enriched prompts were preferred in \textbf{11/20} cases (\textbf{55\%}), and those from \texttt{Llama}-enriched prompts in \textbf{9/20} (\textbf{45\%}). Given the small sample and near parity, we interpret this as \emph{comparable} performance; Figure \ref{fig:llama} represents one example out of study indicating comparable performance.

\paragraph{Takeaway.}
Within this controlled setting, open-source and closed-source LLMs yield \emph{comparable} enrichment quality, supporting reproducible, low-cost adoption. We view this as complementary to our main findings rather than a replacement for them.

\subsection{Additional Analysis}

\begin{figure*}[h]
\centering
\includegraphics[width=\linewidth]{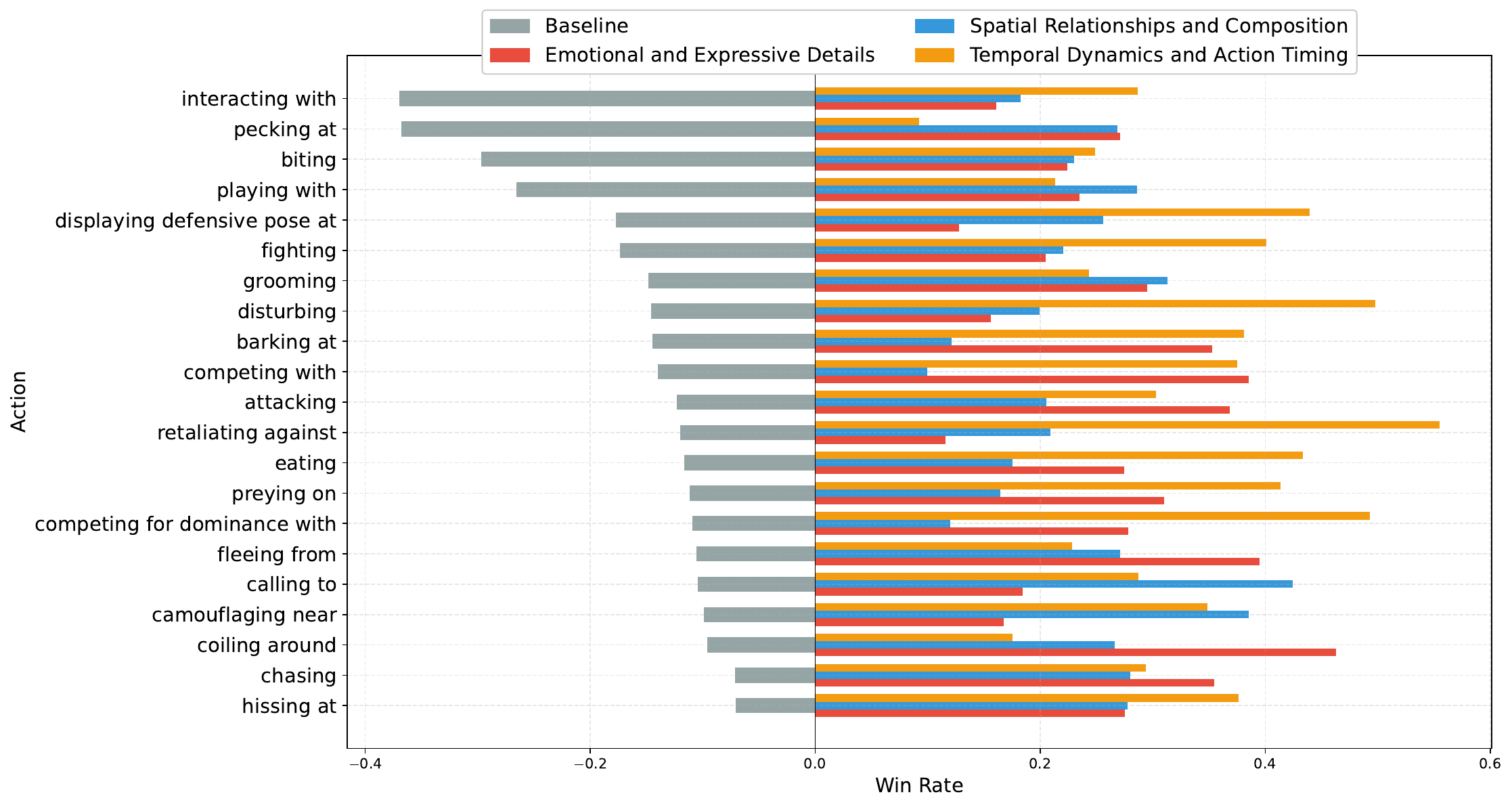}
\caption{Diverged bar graph comparing win rates of baseline and dimension-enriched prompts across different action categories.}
\label{fig:diverged_bar}
\end{figure*}

Figure~\ref{fig:diverged_bar} shows a diverged bar graph comparing baseline prompts versus dimension-enriched prompts across various action categories. This visualization illustrates how each enrichment dimension shifts performance relative to the baseline.

\section{Annotation Details}
\label{subsec:annot_details}
\noindent\textbf{Annotator Instructions:} 3 independent annotators evaluated each generated image by answering: “Does the image truly represent the action in the prompt?” Annotators considered correctness of the entities, plausibility of the depicted action, and subtle cues like emotions, spatial arrangement, and implied motion. They were encouraged to look beyond surface-level accuracy and assess whether the scene convincingly captured the intended relationships and dynamics.

\noindent\textbf{Annotator Details:} We crowdsourced on Amazon Mechanical Turk, 25 annotators in total completed the blind review.

\noindent\textbf{Privacy and Ethics:} Our dataset involves animal subjects with no personal data. The Animal Kingdom dataset and generated images are free of sensitive human information, ensuring compliance with ethical research guidelines and no privacy concerns.

\section{Implementation Details}

We used publicly available model checkpoints and default parameters for image generation. Each prompt was rendered with four random seeds per model. Hyperparameters such as guidance scale, sampling steps, and resolution were kept consistent across models and conditions.

For enrichment, we employed GPT-4 with fixed temperature and token limits to ensure consistent output quality. Minor adjustments were made to each enriched prompt until it provided clear semantic guidance without altering the core meaning of the original prompt.

%% file: tables/axis_rules.tex
\newcolumntype{Y}{>{\RaggedRight\arraybackslash}X} 
\renewcommand{\arraystretch}{1.2} 

\begin{table*}[t]
\centering
\footnotesize
\setlength{\tabcolsep}{3pt}
\begin{tabularx}{\textwidth}{p{0.15\textwidth} p{0.20\textwidth} Y Y}
\toprule
\textbf{Axis} & \textbf{Definition} & \textbf{Category rules \& verb lists} & \textbf{Example prompt} \\
\midrule
Interaction rarity &
Real-world frequency of the (subject A, verb, subject B) interaction. &
\textbf{Frequent:} routine behaviour (e.g., farming, urban)—thousands of sightings.\par
\textbf{Rare:} biologically plausible but only occasionally witnessed.\par
\textbf{Very rare:} geographically improbable or predator–prey reversal. &
\textbf{Frequent:} a Snake attacking a Possum\par
\textbf{Rare:} a Moose attacking a Duck\par
\textbf{Very rare:} an Iguana eating a Cougar \\
\addlinespace[4pt]
Emotional valence &
Dominant affect implied by the verb phrase. &
\textbf{Aggressive:} \{attacking, chasing, biting, eating, competing, fighting, disturbing, retaliating, coiling around, preying on, pecking at\}.\par
\textbf{Defensive:} \{fleeing from, camouflaging near, displaying defensive pose at\}.\par
\textbf{Affiliative:} \{playing with, grooming, interacting with\}.\par
\textbf{Communicative:} \{barking at, hissing at, calling to\}. &
\textbf{Aggressive:} a Camel eating a Swan\par
\textbf{Defensive:} a Fox fleeing from a Cougar\par
\textbf{Affiliative:} a Rat playing with a Possum\par
\textbf{Communicative:} a Dog barking at a Cat \\
\addlinespace[4pt]
Spatial topology &
Coarse physical relation between agents. &
\textbf{Proximal-contact:} physical touch (all aggressive/affiliative verbs except chasing/fleeing).\par
\textbf{Pursuit/Avoidance:} \{chasing, fleeing from\}.\par
\textbf{Distal-non-contact:} vocal/visual signals at a distance. &
\textbf{Proximal-contact:} a Moose attacking a Duck\par
\textbf{Pursuit/Avoidance:} a Kangaroo chasing a Raccoon\par
\textbf{Distal-non-contact:} a Dog barking at a Cat \\
\addlinespace[4pt]
Temporal extent &
Whether the action is momentary or ongoing. &
\textbf{Instantaneous:} \{attacking, biting, pecking at, barking at, hissing at, calling to, displaying defensive pose at\}.\par
\textbf{Extended:} all remaining verbs (e.g., chasing, eating, playing with, camouflaging near). &
\textbf{Instantaneous:} a Goose pecking at a Hamster\par
\textbf{Extended:} a Monkey interacting with an Otter \\
\bottomrule
\end{tabularx}
\caption{Axis definitions, category rules, and examples.}
\label{tab:axis-rules}
\end{table*}

%% file: tables/axis_dist.tex
\newcolumntype{P}[1]{>{\RaggedRight\arraybackslash}p{#1}}

\begin{table*}[t]
\centering
\small
\setlength{\tabcolsep}{4pt}
\begin{tabular}{P{0.15\linewidth} P{0.34\linewidth} P{0.40\linewidth}}
\hline
\textbf{Axis} & \textbf{Distribution} & \textbf{Why this mix?} \\
\hline
Interaction rarity &
Frequent 31\% \;·\; Rare 31\% \;·\; Very rare 38\% & Over-representing very-rare events (e.g., `a goose retaliating against a giraffe'') pushes models beyond memorised patterns. \\ Emotional valence & Aggressive 56\% \;·\; Defensive 20\% \;·\; Affiliative 12\% \;·\; Communicative 12\% & Aggressive/defensive scenes trigger the largest failure modes; the other affects remain well covered. \\ Spatial topology & Pursuit/Avoidance 16\% \;·\; Proximal-contact 64\% \;·\; Distal 20\% & Physical-contact scenes are hardest; pursuit and distal still account for $>$\(\,\)1/3 of prompts. \\ Temporal extent & Instantaneous 28\% \;·\; Extended 72\% & Extended actions (e.g., coiling) require temporal reasoning; instantaneous events supply balance. \\
\hline
\end{tabular}
\caption{Prompt distribution across annotation axes and design rationale.}
\label{tab:axis-dist}
\end{table*}

%% file: tables/axis_acc.tex
\begin{table}[t]
\centering
\resizebox{\linewidth}{!}{
\begin{tabular}{@{}llll@{}}
\toprule
\textbf{Axis} & \textbf{Category} & \textbf{Baseline (\%)} & \textbf{Enriched (\%)} \\
\midrule
\emph{Interaction rarity} & Frequent            & 42.7 & 85.5 \\
                          & Rare                & 50.0 & 82.1 \\
                          & Very rare           & 51.1 & 81.7 \\
\midrule
\emph{Emotional valence}  & Communicative       & 64.4 & 92.2 \\
                          & Affiliative         & 58.9 & 88.3 \\
                          & Aggressive          & 49.3 & 82.0 \\
                          & Defensive           & 28.7 & 77.0 \\
\midrule
\emph{Spatial topology}   & Distal (non-contact) & 51.7 & 85.3 \\
                          & Proximal-contact     & 51.4 & 83.3 \\
                          & Pursuit/Avoidance    & 30.8 & 78.8 \\
\midrule
\emph{Temporal extent}    & Instantaneous       & 55.7 & 85.7 \\
                          & Extended            & 45.2 & 81.9 \\
\bottomrule
\end{tabular}
}
\caption{Stable Diffusion 3.5 Large: majority-vote acceptance by axis for \textit{Baseline} vs \textit{Enriched} prompts. Values are percentages.}
\label{tab:axis-acc}
\end{table}

%% file: tables/dino_clip_eval.tex

\begin{table}[t]
\centering
\begin{tabular}{@{}lrr@{}}
\toprule
\textbf{Model Name} & \textbf{CLIPScore} & \textbf{DinoScore} \\ \midrule
Auraflow            & 0.30               & 0.20               \\
FLUX.1-dev          & 0.38               & 0.30               \\
PixArt-$\Sigma$     & 0.38               & 0.24               \\
Playground v2.5     & 0.56               & 0.35               \\
SD 3.5 large        & 0.38               & 0.24               \\ \bottomrule
\end{tabular}
\caption{Alignment evaluation of Automated metrics with Human preferences, where 1 indicates full alignment, and 0 indicates no alignment.}
\label{tab:dino_clip_eval}
\end{table}

\begin{table}[]
\begin{tabular}{rrr}

\end{tabular}
\end{table}

%% file: tables/action_template.tex
\begin{table*}[h]
\centering
\begin{tabular}{p{0.3\linewidth}p{0.7\linewidth}}
\toprule
\textbf{Plausibility Level} & \textbf{Action Templates} \\
\midrule
Highly Plausible Actions & 
{[}reptile|mammal|bird{]} attacking {[}reptile|mammal|bird{]} \\
& {[}mammal{]} chasing {[}mammal|bird|reptile{]} \\
& {[}mammal|reptile{]} eating {[}mammal|bird|reptile{]} \\
& {[}mammal|reptile|bird{]} fleeing from {[}mammal|bird|reptile{]} \\
& {[}reptile|mammal|bird{]} competing with {[}reptile|mammal|bird{]} \\
& {[}mammal{]} fighting {[}mammal|reptile{]} \\
& {[}bird{]} fighting {[}bird{]} \\
& {[}mammal|reptile|bird{]} disturbing {[}reptile|mammal|bird{]} \\
& {[}reptile|mammal|bird{]} biting {[}reptile|mammal|bird{]} \\
& {[}mammal{]} playing with {[}mammal{]} \\
& {[}bird{]} competing for dominance with {[}bird{]} \\
& {[}mammal|bird{]} grooming {[}mammal|bird{]} \\
& {[}mammal{]} retaliating against {[}reptile|mammal|bird{]} \\
\midrule
Moderately Plausible Actions & 
{[}mammal{]} barking at {[}mammal|reptile{]} \\
& {[}reptile{]} hissing at {[}mammal|bird{]} \\
& {[}reptile|mammal{]} competing for dominance with {[}reptile|mammal{]} \\
& {[}reptile{]} coiling around {[}mammal|bird|reptile{]} \\
& {[}reptile{]} preying on {[}mammal|bird{]} \\
& {[}bird|mammal{]} calling to {[}bird|mammal{]} \\
& {[}bird|mammal{]} fleeing from {[}mammal|bird{]} \\
& {[}reptile{]} camouflaging near {[}mammal|bird{]} \\
\midrule
Less Plausible Actions & 
{[}bird{]} pecking at {[}reptile|mammal{]} \\
& {[}mammal|bird{]} fleeing from {[}reptile{]} \\
& {[}reptile|mammal|bird{]} interacting with {[}mammal|bird|reptile{]} \\
& {[}reptile|mammal{]} displaying defensive pose at {[}reptile|mammal|bird{]} \\
\bottomrule
\end{tabular}
\caption{Action templates grouped by plausibility. These templates guided prompt creation, ensuring diverse scenarios from simple to highly complex and context-dependent.}
\label{tab:action_templates}
\end{table*}